\documentclass{article}

\usepackage{arxiv}
\usepackage[utf8]{inputenc} 
\usepackage[T1]{fontenc}    
\usepackage{hyperref}       
\usepackage{url}            
\usepackage{booktabs}       
\usepackage{amsfonts}       
\usepackage{amsmath,amssymb,amsthm}
\usepackage{nicefrac}       
\usepackage{microtype}      
\usepackage{cleveref}       
\usepackage{lipsum}         
\usepackage{graphicx}
\usepackage{natbib}
\usepackage{doi}
\usepackage{mathtools}
\usepackage{bm}
\usepackage{commath}
\usepackage{diagbox}
\usepackage{multirow}
\usepackage{adjustbox}
\usepackage{xcolor}
\usepackage{listings}
\usepackage{upgreek}
\usepackage{makecell}
\usepackage{algorithm}
\usepackage{algorithmic}

\theoremstyle{plain}
\newtheorem{theorem}{Theorem}
\newtheorem{proposition}[theorem]{Proposition}

\theoremstyle{definition}

\newtheorem{assumption}[theorem]{Assumption}
\theoremstyle{remark}

\newcommand{\eu}{\mathrm{e}\mkern1mu}
\newcommand{\ramuno}{\mathrm{i}\mkern1mu}
\newcommand{\diff}{\mathop{}\!\mathrm{d}}
\DeclareMathOperator{\sinc}{sinc}

\newlength\savewidth

\definecolor{Gray}{gray}{0.5}

\definecolor{Highlight}{HTML}{39b54a}  

\title{Converting Transformers into DGNNs Form}

\newif\ifuniqueAffiliation
\uniqueAffiliationtrue

\author{
Jie Zhang \\
National Central University, Taiwan\\
\texttt{hazdzz@g.ncu.edu.tw}
\And
Mao-Hsuan Mao \\
National Central University, Taiwan\\
\texttt{mmh.nuss@gmail.com}
\And
Bo-Wei Chiu \\
National Central University, Taiwan\\
\texttt{h23468270@g.ncu.edu.tw}
\And
Min-Te Sun \\
National Central University, Taiwan\\
\texttt{msun@csie.ncu.edu.tw}
}

\renewcommand{\shorttitle}{Converting Transformers into DGNNs Form}

\hypersetup{
pdftitle={Converting Transformers into DGNNs Form},
pdfsubject={cs.LG, cs.CL},
pdfauthor={Jie Zhang, Kuan-Chieh Wang, Bo-Wei Chiu, Min-Te Sun},
pdfkeywords={Transformer, DGNN},
}

\begin{document}
\maketitle

\begin{abstract}
Recent advances in deep learning have established Transformer architectures as the predominant modeling paradigm. Central to the success of Transformers is the self-attention mechanism, which scores the similarity between query and key matrices to modulate a value matrix. This operation bears striking similarities to digraph convolution, prompting an investigation into whether digraph convolution could serve as an alternative to self-attention. In this study, we formalize this concept by introducing a synthetic unitary digraph convolution based on the digraph Fourier transform. The resulting model, which we term Converter, effectively converts a Transformer into a Directed Graph Neural Network (DGNN) form. We have tested Converter on Long-Range Arena benchmark, long document classification, and DNA sequence-based taxonomy classification. Our experimental results demonstrate that Converter achieves superior performance while maintaining computational efficiency and architectural simplicity, which establishes it as a lightweight yet powerful Transformer variant.
\end{abstract}

\section{Introduction}
Through the diligent efforts of researchers, Transformers~\citep{NIPS2017_3f5ee243} have played a crucial role in addressing natural language processing tasks~\citep{devlin-etal-2019-bert} since their inception. At the heart of Transformers is the self-attention mechanism that utilizes the scaled dot-product of a query matrix and a key matrix to generate similarity scores, which guide a value matrix. This enables Transformers to capture long-range dependencies and perform parallel computation. Recently, Transformers have even dominated computer vision~\citep{dosovitskiy2021an} and the biology domain~\citep{10.1145/3388440.3412467}.

Because the softmax function binds a query matrix and a key matrix together to compute attention scores~\citep{NIPS2017_3f5ee243}, the high computational cost of self-attention hinders its ability to handle large datasets. Recently, researchers have focused on developing self-attention alternatives with lower time complexity. Approximating the softmax function via kernel functions is a popular choice~\citep{tsai-etal-2019-transformer,choromanski2021rethinking,zhen2022cosformer}. However, this may cause the attention matrix in each attention layer to become low-rank. In this situation, the expressive capability of Transformers significantly declines~\citep{pmlr-v139-dong21a}.

The necessity of the softmax function in self-attention has been questioned. First, the softmax function does not have sufficient ability to express the true data distribution, which constrains the representational capacity of language models and leads to the softmax bottleneck issue~\citep{yang2018breaking}. Moreover, the softmax function inherently fails at robust reasoning across all inputs due to coefficient dispersion with increasing input elements~\citep{velickovic2024softmax}.

Therefore, replacing self-attention has become another well-known option~\citep{pmlr-v139-tay21a,lee-thorp-etal-2022-fnet,9878955}. However, achieving high-rank or full-rank attention matrices while maintaining a time complexity lower than quadratic is a challenge. We address this challenge via synthetic digraph convolution~\footnote{In this work, directed graphs are abbreviated as digraphs.}. Since self-attention is closely related to digraph convolution~\citep{NEURIPS2020_c8512d14}, why not replace self-attention with digraph convolution? This work is based on this hypothesis. We synthesize a unitary digraph convolution called Synvolution, which achieves high performance while maintaining linearithmic time complexity for long sequences. We also apply the kernel polynomial method~\citep{10.1142/S0129183194000842,PhysRevB.49.10154,PhysRevLett.73.1039,Vijay2004,RevModPhys.78.275,Weiße2008} as an alternative technique for the multi-head operation. We refer to Synvolution with the kernel polynomial method as Kernelution. Supported by the theoretical foundation of the kernel polynomial method, the filter of Kernelution can function as any corresponding unitary filter required by distinct datasets.

In this work, we introduce \textbf{Converter}, a Transformer that is converted into a DGNN form. We have evaluated Converter on Long-Range Arena benchmark~\citep{tay2021long}, long document classification~\citep{9878955}, and DNA sequence-based taxonomy classification~\citep{9878955}. The experimental results demonstrate that Converter outperforms previous Transformer variants. This demonstrates that Converter is a lightweight, efficient, and powerful neural network. Our key contributions are summarized as follows:
\begin{itemize}
\item We propose Synvolution, a novel self-attention alternative with linearithmic time complexity.
\item We apply the kernel polynomial method as an alternative to the multi-head operation, and propose kernel polynomial loss to simulate a dynamic kernel. We name Synvolution with the kernel polynomial method as Kernelution.
\item We propose Gated Feed-Forward Networks in place of the vanilla Feed-Forward Networks for complex-valued input and real-valued output.
\item We apply PostNorm with ScaleNorm for both real-valued and complex-valued tensor.
\end{itemize}
\section{Related Work}
\noindent\textbf{Self-Attention Alternatives.}
Central to self-attention is the scaled dot-product operation performed on a pair of query and key matrices, yielding an affinity matrix with a softmax function. Due to its high time complexity, various alternative methodologies have been proposed to approximate or replace scaled dot-product attention. Approximate approaches typically involve sparse~\citep{child2019generating,Kitaev2020Reformer,NEURIPS2020_c8512d14}, low-rank~\citep{choromanski2021rethinking}, or sparse $\text{+}$ low-rank~\citep{NEURIPS2021_9185f3ec} techniques. For replacements, common approaches include convolution~\citep{yu2022metaformer}, pooling~\citep{Yu_2022_CVPR}, and discrete Fourier transform~\citep{lee-thorp-etal-2022-fnet}. Recently, a spatial construction method~\citep{pmlr-v139-tay21a,9878955} has emerged, utilizing the inverse process of matrix decomposition to synthesize attention matrices. This approach has inspired us to propose a synthetic attention.

\noindent\textbf{Multi-Head Attention.}
Multi-head attention may not be more efficient than single-head. \citet{NEURIPS2019_2c601ad9} discover that over 50\% attention heads can be pruned during the testing phase. Similarly, \citet{voita-etal-2019-analyzing} conclude that only a small subset of heads is critical for translation tasks. Furthermore, \citet{cordonnier2020multihead} identify redundant feature representations in multi-head self-attention. Additionally, \citet{pmlr-v119-bhojanapalli20a} observe that a large number of heads cause a low-rank bottleneck, which restricts the representation capacity of Transformers. Based on these observations, we focus on single-head attention.

\noindent\textbf{Digraph Fourier Transform.}
The Digraph Fourier Transform serves as the cornerstone for digraph convolution, based on the fundamental assumption that it requires a diagonalizable digraph shift operator~\citep{10388222}. There are two distinct categories of methods that aim to achieve the digraph Fourier transform. The first category involves replacing the eigendecomposition with an alternative matrix decomposition to build orthogonal or unitary bases~\citep{6409473,6638850,6808520,7746675}, while the second entails spatially constructing a normal digraph shift operator~\citep{Chung2005,PhysRevE.95.022302,FANUEL2018189}. We draw inspiration from the two categories of methods and propose a unique one that synergizes both approaches.

\noindent\textbf{Structured Matrices.}
In random and linear projections, structured matrices are aimed to reduce time complexity. One major effect in random projection is improving Johnson-Lindenstrauss transform~\citep{10.1145/1132516.1132597,10.1145/1806689.1806737}. Another prominent effect of structured matrices is random features~\citep{pmlr-v28-le13,NIPS2016_53adaf49}. By replacing random entities with learnable parameters, linear projection has been in developed recent years. A well-known example is the Adaptive Fastfood transform~\citep{Yang_2015_ICCV}, which extends the vanilla Fastfood transform~\citep{pmlr-v28-le13} by incorporating learnable diagonal matrices. Building on these developments, \citet{moczulski2016acdc} unify these structured matrices under the SELL matrix family and introduce ACDC and AFDF matrices. These approaches enlighten us to design structured unitary matrices under quadratic time complexity.
\section{Background and Preliminary}
\subsection{Self-Attention and Multi-Head Self-Attention}
Given an input signal $\mathbf{X} \in \mathbb{R}^{N \times D}$, the self-attention (SA)~\citep{NIPS2017_3f5ee243} is defined as
\begin{equation}\label{eq:shsa}
\centering
\mathrm{SA}(\mathbf{X}) = 
\mathrm{Softmax}\left(\frac{\mathbf{X}\mathbf{W}_{\text{Q}}(\mathbf{X}\mathbf{W}_{\text{K}})^{\mathrm{T}}}{\tau}\right)\mathbf{X}\mathbf{W}_{\text{V}},
\end{equation}
where $\tau=\sqrt{D_{h}}$ is the temperature parameter, the softmax function is applied row-wise, $\mathrm{T}$ represents the transpose operation, $\mathbf{X}\mathbf{W}_{\text{Q}} \in \mathbb{R}^{N \times D_{h}}$, $\mathbf{X}\mathbf{W}_{\text{K}} \in \mathbb{R}^{N \times D_{h}}$, and $\mathbf{X}\mathbf{W}_{\text{V}} \in \mathbb{R}^{N \times D_{h}}$ are referred to as the query, key, and value matrix, in which $\mathbf{W}_{\text{Q}} \in \mathbb{R}^{D \times D_{h}}$, $\mathbf{W}_{\text{K}} \in \mathbb{R}^{D \times D_{h}}$, and $\mathbf{W}_{\text{V}} \in \mathbb{R}^{D \times D_{h}}$ are learnable parameters. Conventionally, the Multi-Head Self-Attention (MHSA) with $H$ heads is commonly chosen to enhance performance. It is defined as
\begin{equation}\label{eq:mha}
\centering
\mathrm{MHSA}(\mathbf{X}) = \left(\Big{\Vert}_{h=1}^{H}\mathrm{SA}(\mathbf{X})_{h}\right)\mathbf{W}_{\text{O}},
\end{equation}
where $\Big{\Vert}$ represents the concatenation operation, and $\mathbf{W}_{\text{O}} \in \mathbb{R}^{HD_{h} \times D}$ is a learnable parameter. Here, $D_{h} = D/H$.

\subsection{Digraph Signal Processing}
In this work, we follow the definitions of digraph signal processing (DGSP)~\citep{10388222,6409473,6638850,6808520}. A digraph is represented as $G = \{\mathcal{V},\mathcal{E}\}$, where $\mathcal{V}$ denotes the set of vertices with $|\mathcal{V}| = N$, and $\mathcal{E} \subseteq \mathcal{V}\times\mathcal{V}$ represents the set of edges. A digraph adjacency matrix is denoted by $\mathbf{A} \in \mathbb{C}^{N \times N}$, where each element corresponds to an edge, and the module of the weight signifies the degree of the edge.

In DGSP, a digraph shift operator (DGSO) $\mathbf{S}_{G}$ is a matrix defining the manner in which a digraph signal transitions from one node to its neighboring nodes based on the underlying digraph topology. More precisely, a DGSO constitutes a local operator that substitutes the digraph signal value at each node with a linear combination of its neighboring nodes' values. It is a fundamental assumption to employ a diagonalizable (normalized) digraph adjacency or Laplacian matrix as a DGSO to perform a digraph convolution. In this situation, every digraph signal can be represented as a linear combination of the eigenvectors of the DGSO. 

A digraph filter $\mathcal{H}_{\mathbf{\theta}}(\mathbf{S}_{G}) \in \mathbb{C}^{N \times N}$ is a function of the DGSO termed the digraph frequency response function where eigenvalues are perceived as digraph frequencies. In DGSP, a kind of widely adopted digraph filter is based on polynomials, such a digraph filter is defined as $\mathcal{H}_{\mathbf{\theta}}(\mathbf{S}_{G}) = \sum^{K}_{k=0}{{\theta}_{k}\mathbf{S}_{G}^{k}}$, where ${\theta}_{k}$ is the corresponding coefficient. This form of digraph filter is called the Finite Impulse Response (FIR) filter. A typical FIR filter is based on the Chebyshev polynomials of the first kind~\citep{HAMMOND2011129}~\footnote{The following is abbreviated as Chebyshev polynomials.}. For input $x \in [-1, 1]$, through three-term recurrence relations, the Chebyshev polynomials are obtained as $T_{k}(x) = 2x\cdot{T}_{k-1}(x) - T_{k-2}(x)$, with $T_{0}(x) = 1$ and $T_{1}(x) = x$. Combining a polynomial based filter, the procedure of the digraph convolution (DGConv) can be articulated as
\begin{equation}\label{eq:digraph_conv_eigenvalue}
\centering
\mathrm{DGConv}(\mathbf{S}_{G}, \mathbf{X}) 
= \mathbf{U}^{-1}\left[\mathcal{H}_{\mathbf{\theta}}(\mathbf{\Lambda})\odot(\mathbf{U}\mathbf{X})\right],
\end{equation}
where $\mathbf{U} \in \mathbb{C}^{N \times N}$ is the eigenvector matrix of $\mathbf{S}_{G}$. $\hat{\mathbf{X}} = \mathbf{U}\mathbf{X} \in \mathbb{C}^{N \times D}$ represents the digraph Fourier transform applied to the input signals, while $\mathbf{X} = \mathbf{U}^{-1}\hat{\mathbf{X}} \in \mathbb{C}^{N \times D}$ denotes the inverse transform.

\section{Proposed Method}
\begin{figure}[t]
\begin{center}
\includegraphics[scale=0.4]{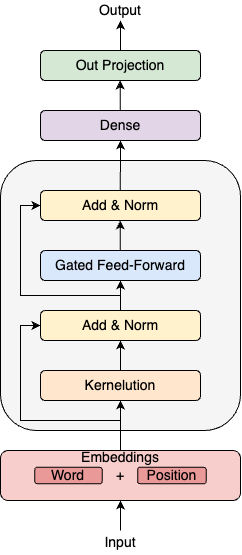}
\end{center}
\caption{Converter architecture.}
\label{fig:converter_architecture}
\end{figure}

\subsection{Synvolution}
Let $\mathbf{A} = \mathbf{X}\mathbf{W}_{\text{Q}}(\mathbf{X}\mathbf{W}_{\text{K}})^{\mathrm{T}}/{\tau} \in \mathbb{R}^{n \times n}$ be an affinity matrix, the attention matrix $\mathcal{A} = \mathrm{softmax}(\mathbf{A})$ in Equation~\ref{eq:shsa} can be reformulated as a right stochastic normalized affinity form $\mathrm{SA}(\mathbf{X}) = \widetilde{\mathbf{D}}^{-1}\widetilde{\mathbf{A}}\mathbf{X}\mathbf{W}_{\text{V}}$, where $\widetilde{\mathbf{A}} = \exp(\mathbf{A}) \in \mathbb{R}^{n \times n}$ is defined as the affinity matrix after the element-wise exponentiation operation $\exp(\cdot)$, and $\widetilde{\mathbf{D}}_{u,u} = \sum_{v}\widetilde{\mathbf{A}}_{u,v} \in \mathbb{R}^{n \times n}$ is the corresponding degree matrix. When treating $\widetilde{\mathbf{A}}$ as a digraph adjacency matrix, we found that self-attention closely resembles digraph convolution. First, each element in either an attention matrix or a DGSO can be considered as a similarity from source entity to target entity. Second, both self-attention and digraph convolution can be degenerated to graph convolution form. For self-attention, this occurs when the query matrix is equal to the key matrix in each head, resulting in unidirectional symmetric self-attention. Similarly, for digraph convolution, the achievement of graph convolution can be implemented by symmetrizing the adjacency matrix of a digraph. Third, the softmax function in self-attention results in a row-wise normalized digraph adjacency form. 

Since digraph convolution closely resembles self-attention, we investigated replacing self-attention with digraph convolution. Under this hypothesis, a Transformer can be converted into a DGNN form. Based on this insight, we propose Converter. In this work, we decide to construct the DGSO directly. We develop a learnable unitary matrix as a DGSO through the inverse process of eigendecomposition. Our method consists of two phases. In the first phase, we synthesize the required eigenvalues through the following process.
\begin{equation}\label{eq:eigenvalue_gap}
\centering
\eu^{\ramuno\mathbf{\Lambda}}
= \exp\left[\ramuno\cdot\mathrm{diag}\left(\mathrm{pool}_{\text{avg}}\left[\mathrm{SIREN}(\mathbf{X})\right]\right)\right].
\end{equation}
Here, $\mathrm{SIREN}$ represents a 2-layer MLP with the sine function~\citep{NEURIPS2020_53c04118}, $\mathrm{pool}_{\text{avg}}(\cdot)$ is a 1D global average pooling, and $\mathrm{diag}(\cdot)$ is a diagonalize operation. We adopt the sine function because it demonstrates a remarkable ability in signal processing~\citep{NEURIPS2020_53c04118}.

In the second phase, we focus on constructing the necessary unitary eigenvector matrix through the inverse process of LQ factorization. Based on the Givens rotation method~\citep{doi:10.1137/0106004}, an arbitrary square matrix $\bm{\Phi} \in \mathbb{C}^{N \times N}$ can be decomposed into a product of a lower triangular matrix and Givens rotation matrices. Hence, we have
\begin{equation}\label{eq:givens_lq}
\centering
\bm{\Phi} 
= \mathbf{L}\mathbf{Q} 
= \mathbf{L}\left(\prod_{j=N}^{2}\prod_{i=j-1}^{1}\mathbf{G}_{i,j}\right),
\end{equation}
where $\mathbf{L} \in \mathbb{C}^{N \times N}$ is a lower triangular matrix, and $\mathbf{G}_{i,j} \in \mathbb{C}^{N \times N}$ is a Givens rotation matrix that resembles an identity matrix with the exception of the elements
\begin{equation}\label{eq:givens_rotation_matrix}
\centering
\begin{bmatrix}
G_{ii} & G_{ij} \\
G_{ji} & G_{jj}
\end{bmatrix} =
\begin{bmatrix}
\overline{c} & -s \\
\overline{s} & c
\end{bmatrix} =
\begin{bmatrix}
\eu^{-\ramuno(\frac{\alpha+\beta}{2})}\cos{(\frac{\gamma}{2})} 
& -\eu^{\ramuno(\frac{\alpha-\beta}{2})}\sin{(\frac{\gamma}{2})} \\
\eu^{-\ramuno(\frac{\alpha-\beta}{2})}\sin{(\frac{\gamma}{2})} 
& \eu^{\ramuno(\frac{\alpha+\beta}{2})}\cos{(\frac{\gamma}{2})}
\end{bmatrix},
\end{equation}
which characterized by parameters $\alpha$, $\beta$, and $\gamma \in [0, 2\pi]$. This methodology necessitates ${\left(N(N-1)\right)}/{2}$ pairs of Givens rotation matrices, i.e., it requires $\mathcal{O}(N^{2})$ space complexity. By reorganizing Givens rotation matrices, inserting permutation matrices, and repeating the patten, we have
\begin{equation}\label{eq:givens_lhhp}
\centering
\begin{split}
\bm{\Phi} 
&= \mathbf{L}\left(\prod_{l=1}^{L}\left(\prod_{i=N-1}^{1}\mathbf{G}^{(l)}_{i,i+1}\right)\left(\prod_{j=1}^{N-1}\mathbf{G}^{(l)}_{j,j+1}\right)\mathbf{P}^{(l)}\right)\\
&= \mathbf{L}\left(\prod_{l=1}^{L}\mathbf{H}^{(l)}_{\text{l}}\mathbf{H}^{(l)}_{\text{u}}\mathbf{P}^{(l)}\right).
\end{split}
\end{equation}
Here, $\mathbf{H}^{(l)}_{\text{l}} \in \mathbb{C}^{N \times N}$ is a lower unitary Hessenberg matrix, $\mathbf{H}^{(l)}_{\text{u}} \in \mathbb{C}^{N \times N}$ is an upper unitary Hessenberg matrix, and $\mathbf{P}^{(l)} \in \mathbb{R}^{N \times N}$ is a permutation matrix that either learnable~\citep{mena2018learning}, fixed~\citep{pmlr-v162-dao22a}, or even an identity matrix $\mathbf{I}_{N} \in \mathbb{R}^{N \times N}$.

We refer to Equation~\ref{eq:givens_lhhp} as the order-$L$ LHHP parametrization, $L$-LHHP for short, denoted by $\bm{\Phi}_{L-\text{LHHP}}$. In particular, when the lower triangular matrix $\mathbf{L}$ degenerates to a diagonal matrix $\mathbf{D}$, we term this pattern the order-$L$ DHHP parametrization, $L$-DHHP for short, denoted by $\bm{\Phi}_{L-\text{DHHP}}$. It requires $2L(N-1)$ pairs of Givens rotation matrices, which means the space complexity is $\mathcal{O}(LN)$. We observed that each unitary factor matrix resulting from the multiplication of lower and upper unitary Hessenberg matrices in the order-$L$ DHHP parametrization is dense rather than sparse, unlike the schemes proposed in \citep{KHALITOV2022160}. Since our method is based on the Givens rotation method, we make Assumption~\ref{asmp:dhhp_givens_num_upper_bound}. Under this assumption, we can establish the following propositions.

\begin{assumption}\label{asmp:dhhp_givens_num_upper_bound}
For constructing an arbitrary $N \times N$ dense unitary matrix, at most $\lceil\frac{N}{4}\rceil$ orders are sufficient for $L$-DHHP.
\end{assumption}

\begin{proposition}\label{prop:dhhp_universal}
$L$-DHHP captures the discrete unitary transforms, including discrete Fourier transform (DFT), the discrete Walsh–Hadamard transform (DWHT), the discrete cosine transform (DCT), the discrete sine transform (DST), and their inverses exactly.
\end{proposition}

\begin{proposition}\label{prop:dhhp_time_complexity}
Given an input signal $\mathbf{x} \in \mathbb{C}^{N}$ and an output signal $\mathbf{y} \in \mathbb{C}^{N}$, the time complexity of $L$-DHHP as a discrete unitary transform with the fast implementation as $\mathbf{y} = \bm{\Phi}\mathbf{x}$ is $\mathcal{O}(LN\log{N})$.
\end{proposition}

\begin{proposition}\label{prop:dhhp_full_rank}
$L$-DHHP is full-rank if and only if the diagonal matrix $\mathbf{D}$ is unitary.
\end{proposition}

We refer to this self-attention alternative as Synvolution:
\begin{equation}\label{eq:synvolution}
\centering
\mathrm{Synv}(\mathbf{X}\mathbf{W}_{\text{V}})
= \bm{\Phi}^{-1}\left[\exp(\ramuno\mathbf{\Lambda})\odot(\bm{\Phi}\mathbf{X}\mathbf{W}_{\text{V}})\right],
\end{equation}
where $\mathbf{X}\mathbf{W}_{\text{V}} \in \mathbb{C}^{N \times D}$ is denoted as the value matrix. Unlike FFT-based convolution~\citep{mathieu2013fast}, where the discrete unitary matrix is fixed and data-independent, the required parameters in $L$-DHHP are learnable and data-dependent. We adopt a similar processing method to that described in Equation~\ref{eq:eigenvalue_gap} to obtain the synthetic eigenvector matrix. For convenience, we set $L=1$, $\mathbf{P}^{(1)} = \mathbf{I}_{N}$, and $\mathbf{D}$ is unitary to obtain a dense unitary matrix that serves as the desired unitary eigenvector matrix. More details about the fast implementation of $1$-DHHP as a discrete unitary transform are provided in the appendix.


\subsection{Kernelution}

\begin{figure*}[t]
\begin{center}
\includegraphics[scale=0.4]{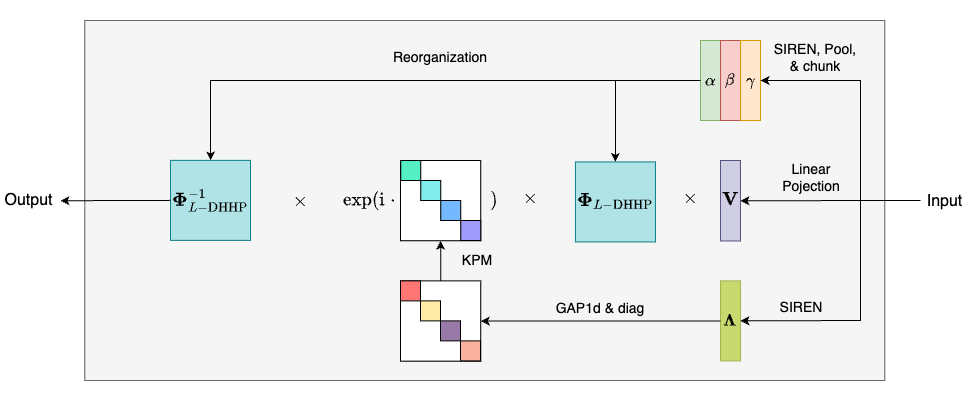}
\end{center}
\caption{Illustration of the entire Kernelution process.}
\label{fig:kernelution}
\end{figure*}

\subsubsection{Chebyshev Polynomial Interpolation}
The multi-head operation, a common approach to enhance performance in Transformers, lacks solid theoretical support. In the contrast, FIR filters have a theoretical support in spectral graph theory~\citep{chung1997spectral}. Let $f(x)$ be the target function, then our goal is to approximate it with the smallest round-off error. Directly manipulating orthogonal polynomials to filter complex-valued signals is challenging, but using them to represent the argument function of signals is straightforward. To achieve it, we can choose an arbitrary orthogonal polynomial basis such as the Bernstein basis, Jacobi basis (including Chebyshev, Gegenbauer, Legendre, and Zernike bases), or even monomial basis. Consider the Chebyshev basis as an example. Given an arbitrary continuous function $f(x) \in C([-1,1])$ and a truncated Chebyshev polynomial $p$ with $K$ orders, then the target function $f(x)$ can be approximated as
\begin{equation}\label{eq:cpi}
\centering
f(x) \approx p(x) = \frac{1}{2}{\mu}_{0} + \sum_{k=1}^{K}{\mu}_{k}{T}_{k}(x),
\end{equation}
where $\mu_{k} \approx \frac{2}{K+1}\sum_{j=0}^{K}f(x_{j})T_{k}(x_{j})$ is the Chebyshev coefficient, and $x_{j}$ is the sampling Chebyshev node. This technique is termed the Chebyshev polynomial interpolation (CPI)~\citep{10.1137/1.9781611975949}. The operation on Chebyshev polynomial interpolation is considerably straightforward since Chebyshev polynomials are isomorphic with Fourier series. For differentiable or analytic functions, we have the following theorems.

\begin{theorem}[CPI for differentiable functions~\citep{10.1137/1.9781611975949}]\label{theorem:cpi_diff}
Let $\upsilon \geq 0$ and $\kappa > \upsilon$ be integers. Consider a function $f(x)$ whose derivatives up to order $\upsilon-1$ are absolutely continuous on $[-1, 1]$, and suppose $\lVert{\frac{\diff^{\upsilon}}{\diff{x}^{\upsilon}} f(x)}\rVert_{1} = \Upsilon$. For the $\kappa$-th degree Chebyshev interpolant $p(x)$, the following bounds hold: (1)~$\lVert{\mu_{\kappa}}\rVert \leq \frac{2\Upsilon}{\pi(\kappa-\upsilon)^{\upsilon+1}}$. (2)~$\lVert{f(x) - p(x)}\rVert \leq \frac{4\Upsilon}{\pi\upsilon(\kappa-\nu)^{\upsilon}}$.
\end{theorem}

\begin{theorem}[CPI for analytic functions~\citep{10.1137/1.9781611975949}]\label{theorem:cpi_ana}
Let $\kappa \geq 1$ be an integer and $f(x)$ an analytic function on $[-1, 1]$ that extends analytically to the open Bernstein ellipse $E_{\rho}$ with $\lVert{f(x)}\rVert \leq M$ for some $M$. For the $\kappa$-th degree Chebyshev interpolant $p(x)$, the following bounds hold: (1)~$\lVert{\mu_{0}}\rVert \leq M$. (2)~$\lVert{\mu_{\kappa}}\rVert \leq 2M\rho^{-\kappa}$. (3)~$\lVert{f(x)-p(x)}\rVert \leq \frac{4M\rho^{-\kappa}}{\rho-1}$.
\end{theorem}

Both Theorem~\ref{theorem:cpi_diff} and Theorem~\ref{theorem:cpi_ana} tell us that we can utilize the Chebyshev polynomial filter to approximate any continuous target function that lies in the range of $C[-1, 1]$ with a small round-off error.

\subsubsection{Kernel Polynomial Method}
In reality, the target function is probably discontinuous or singular in the polynomial interpolation interval. In this situation, the accuracy of the Chebyshev polynomial interpolation reduces to $\mathcal{O}(1)$ near discontinuities or singularities. Sufficiently far away from discontinuities or singularities, the convergence will be slowed to $\mathcal{O}(K^{-1})$. During the approximation process, oscillations will be present near discontinuities or singularities and they will not diminish as $K \to \infty$. This type of oscillation is termed the Gibbs oscillation, and this situation is known as the Gibbs phenomenon~\citep{Hewitt1979}.

To mitigate Gibbs oscillations, we apply a Gibbs damping factor $g_{k}$, which represented as a function of $\frac{k}{K+1}$, to each term of the Chebyshev polynomials. For any $f(x)$, we have
\begin{equation}\label{eq:kpm}
\centering
{f(x)}\approx{p}_{\text{KP}}(x) = \frac{1}{2}g_{0}\mu_{0} + \sum_{k=1}^{K}g_{k}\mu_{k}T_{k}(x).
\end{equation}
This modification of the Chebyshev coefficients is equivalent to the convolution of $p(x)$ with a kernel $\mathcal{K}(x,x_{0}) = \frac{2}{\pi\sqrt{1-x^{2}}}\left(\frac{1}{2}g_{0} + \sum_{k=1}^{K}g_{k}T_{k}(x)T_{k}(x_{0})\right)$ that $p_{\text{KP}}(x) = \int_{-1}^{1}\mathcal{K}(x,x_{0})f(x_{0})\diff{x_{0}}$. Thus, this method is also called the kernel polynomial method. It is widely employed in computational physics for calculating the density of states and other spectral properties of large quantum systems. 

Gibbs damping factors are a family of coefficients that satisfy three conditions: (1)~$g_{k} > 0$. (2)~$g_{0} = 1$. (3)~$\lim_{K \to \infty} {g_{1} \to 1}$. The conditions (1) and (2) are particularly valuable in real-world applications~\citep{RevModPhys.78.275,Weiße2008}. The first condition ensures that approximations of positive quantities remain positive, while the second conserves the integral of the expanded function $\int_{-1}^{1}p_{\text{KPM}}(x)\diff{x} = \int_{-1}^{1}f(x)\diff{x}$. Notably, $g_{k} = 1$ is the simplest Gibbs damping factor attributed to the Dirichlet kernel. More details about Gibbs damping factors are in the appendix.

Clearly, finding an appropriate kernel is crucial for approximation, as it determines whether the round-off error is minimized or not. As indicated in \citep{RevModPhys.78.275,Weiße2008}, kernel choices are data-dependent. More specifically, given a target function, we need to match an appropriate kernel and manually tune its hyperparameters (if the kernel has any) based on experience. Since the target function is unknown, we relax each $\mu_{k}$ with a learnable parameter $w_{k}$. The effectiveness of the Gibbs damping factors lie in their ability to reduce the weight of each term of the Chebyshev coefficients, thereby mitigating the contributions of higher-order terms. Based on this observation, and in order to prevent over-fitting, we propose the following loss function which is named the kernel polynomial loss (KPL):
\begin{equation}\label{eq:kpl}
\centering
\mathcal{L}_{\text{KP}} = \int_{-1}^{1}\abs{\frac{\diff{f(x)}}{\diff{x}}}^{2}\diff{x} \approx \sum_{k=1}^{K}{\pi}{k}^{2}\abs{w_{k}}^{2}.
\end{equation}
This results in an intuitive penalty applied to the Chebyshev coefficients, with higher order Chebyshev coefficients incurring greater penalties than the lower ones. It causes the Chebyshev polynomial interpolation with the kernel polynomial loss to simulate the kernel polynomial method with a learnable kernel. We apply the kernel polynomial method with Synolution, which turns out what we call Kernelution. The corresponding formula is defined as
\begin{equation}\label{eq:kernelution}
\centering
\mathrm{Kern}(\mathbf{X}\mathbf{W}_{\text{V}})
= \bm{\Phi}^{-1}\left[\exp\left(\ramuno\cdot{p}_{\text{KP}}(\mathbf{\Lambda})\right)\odot(\bm{\Phi}\mathbf{X}\mathbf{W}_{\text{V}})\right].
\end{equation}
It is worth noting that the kernel polynomial method is not the only operation compatible with Synvolution. Depending on practical requirements, Synvolution can also be made compatible with the multi-head operation, similar to other attention mechanisms. This means Synvolution can be equipped as a substitute for self-attention in Transformer-based models.

\subsection{Gated Feed-Forward Network and PostScaleNorm}
Both Synvolution and Kernelution effectively represent the direction and model the relationship between feature tokens in the spectral domain. A tricky problem is that the output of either Synvolution or Kernelution is complex-valued, whereas the labels are real-valued. This conflict motivates us to design a layer that maps a complex-valued tensor into a real-valued tensor. We propose a Gated Feed-Forward Network (GFNN) to solve this issue.
\begin{equation}\label{eq:gffn}
\centering
\mathrm{GFFN}(\mathbf{X}) = \left[\mathrm{softplus}(\Re(\mathbf{X})\mathbf{W}_{\Re})\odot\tanh(\Im(\mathbf{X})\mathbf{W}_{\Im})\right]\mathbf{W}_{\text{O}},
\end{equation}
where $\mathbf{W}_{\Re} \in \mathbb{R}^{D \times D_\text{hid}}$, $\mathbf{W}_{\Im} \in \mathbb{R}^{D \times D_\text{hid}}$ and $\mathbf{W}_{\text{O}} \in \mathbb{R}^{D_\text{hid} \times D}$ are trainable weight matrices. We let the real part to learn the magnitude, and the imaginary part to learn the sign. Besides, we apply the PostNorm architecture~\citep{wang-etal-2019-learning-deep} with ScaleNorm~\citep{nguyen-salazar-2019-transformers} across the whole model, namely PostScaleNorm. Specifically, we apply $\mathrm{ScaleNorm}(\mathbf{Z} + \zeta \cdot \Re(\mathbf{Z}) + (1 - \zeta) \cdot \Im(\mathbf{Z}))$ for a complex-valued signal $\mathbf{Z}$, where $\zeta \in [0, 1]$ is a learnable parameter.
\section{Experiments}
In this section, we test the scalability and performance of Converter in three different domains: (1)~Long-Range Arena benchmark, (2)~long document classification, and (3)~DNA sequence-based taxonomy classification. We conducted all experiments on a NVIDIA DGX-1 equipped with two 20-core Intel Xeon E5-2698 v4 CPUs @ 2.2 GHz, 512 GB of RAM, and 8 NVIDIA Tesla V100 GPUs, each with 16 GB of GPU memory. The code is implemented using PyTorch~\citep{NEURIPS2019_bdbca288}. Following \citet{neishi-yoshinaga-2019-relation}, we adopt a 2-layer GRU for position embedding, which is denoted as RPE. We adopt the AdamW optimizer~\citep{loshchilov2018decoupled} and apply cross-validation to report the best hyperparameters. We apply the following loss functions as the metric to evaluate our model.
\begin{equation}
\centering
\begin{split}
\mathcal{L} = (1-\eta)\cdot\mathcal{L}_{\text{CE}} + \eta\cdot\mathcal{L}_{\text{KP}}
\end{split}
\end{equation}
Here, $\eta \in {[0, 1)}$ is a tunable hyperparameter that needs to be selected manually, and $\mathcal{L}_{\text{CE}}$ denotes cross-entropy loss.

\subsection{Long-Range Arena Benchmark}

\begin{table*}[!ht]
\centering
\caption{Accuracy results (\%) on the Long-Range Arena benchmark. The best result is in bold and the second best is underlined.}
\label{tab:lra_res}
\begin{tabular}{lcccccc}
\toprule
\textbf{Model} & \textbf{ListOps}~$\uparrow$ & \textbf{Text}~$\uparrow$ & \textbf{Retrieval}~$\uparrow$ & \textbf{Image}~$\uparrow$ & \textbf{Pathfinder}~$\uparrow$ & \textbf{Avg.}~$\uparrow$ \\
\midrule
Vanilla Trans.~\citep{NIPS2017_3f5ee243} & 36.37 & 64.27 & 57.46 & 42.44 & 71.40 & 54.39 \\
Sparse Trans.~\citep{child2019generating} & 17.07 & 63.58 & 59.59 & 44.24 & 71.71 & 51.24 \\
Reformer~\citep{Kitaev2020Reformer} & 37.27 & 56.10 & 53.40 & 38.07 & 68.50 & 50.67 \\
Longformer~\citep{beltagy2020longformer} & 35.63 & 62.85 & 56.89 & 42.22 & 69.71 & 53.46 \\
Linformer~\citep{wang2020linformer} & 35.70 & 53.94 & 52.27 & 38.56 & 76.34 & 51.36 \\
BigBird~\citep{NEURIPS2020_c8512d14} & 36.05 & 64.02 & 59.29 & 40.83 & 74.87 & 55.01 \\
Linear Trans.~\citep{pmlr-v119-katharopoulos20a} & 16.13 & 65.90 & 53.09 & 42.34 & 75.30 & 50.55 \\
Sinkhorn Trans.~\citep{pmlr-v119-tay20a} & 33.67 & 61.20 & 53.83 & 41.23 & 67.45 & 51.29 \\
Performer~\citep{choromanski2021rethinking} & 18.01 & 65.40 & 53.82 & 42.77 & 77.05 & 51.41 \\
Synthesizer~\citep{pmlr-v139-tay21a} & 36.99 & 61.68 & 54.67 & 41.61 & 69.45 & 52.88 \\
Nystr{\"{o}}mformer~\citep{xiong2021Nystromformer} & 37.15 & 65.52 & \underline{79.56} & 41.58 & 70.94 & 58.95 \\
Luna-256~\citep{ma2021luna} & 37.98 & 65.78 & \underline{79.56} & \underline{47.86} & 78.55 & \underline{61.95} \\
FNet~\citep{lee-thorp-etal-2022-fnet} & 35.33 & 65.11 & 59.61 & 38.67 & 77.80  & 55.30 \\
cosFormer~\citep{zhen2022cosformer} & 37.90 & 63.41 & 61.36 & 43.17 & 70.33 & 55.23 \\
Paramixer (Chord)~\citep{9878955} & \underline{39.71} & \underline{78.87} & 78.73 & 44.68 & \underline{79.16} & 58.91 \\
\midrule
Converter (ours) & \textbf{60.38} & \textbf{86.44} & \textbf{83.41} & \textbf{61.02} & \textbf{88.43} & \textbf{75.94} \\
\bottomrule
\end{tabular}
\end{table*}

The Long-Range Arena (LRA)~\citep{tay2021long} is a public benchmark established with the aim of evaluating the ability of efficient Transformers to model long-sequence data. This benchmark contains five multi-class classification tasks from distinct domains, including ListOps~\citep{nangia-bowman-2018-listops}, Text~\citep{maas-etal-2011-learning}, Retrieval~\citep{radev-etal-2009-acl}, Image~\citep{krizhevsky2009learning}, and Pathfinder~\citep{NEURIPS2018_ec895663,Kim2020Disentangling}. ListOps consists of digits, operators such as MAX, MEAN, MEDIAN, and SUM\_MOD, and brackets. Each operator in a sequence processes the items in a list and outputs a digit. Text consists of sequences represented at the byte/character-level, which significantly increases its difficulty. In this task, models must classify each review as positive or negative, making it a binary classification task. Retrieval is similar to the Text task with a byte/character-level setting. Image is the CIFAR-10 task~\citep{krizhevsky2009learning} for image classification. The input data consists of sequences of pixels derived from flattening $32 \times 32$ images into a 1D array with the length of 1024. Pathfinder is motivated by cognitive psychology~\citep{Houtkamp2010Parallel}. In this task, a synthetic image measures $32 \times 32$ pixels and features two highlighted endpoints depicted as circles, connected by a dashed path. Each image contains distractor paths, adding complexity. The models must determine whether a dashed path connects the two highlighted endpoints. As in the Image task, the input has to be converted into a sequence with length 1024.

In the interest of ensuring a fair comparison, we follow the experiment settings outlined in \citep{tay2021long} and evaluate Converter on the aforementioned tasks. For baselines, we include the vanilla Transformer~\citep{NIPS2017_3f5ee243} and 14 Transformer variants: Sparse Transformer~\citep{child2019generating}, Reformer~\citep{Kitaev2020Reformer}, Longformer~\citep{beltagy2020longformer}, Linformer~\citep{wang2020linformer}, BigBird~\citep{NEURIPS2020_c8512d14}, Linear Transformer~\citep{pmlr-v119-katharopoulos20a}, Sinkhorn Transformer~\citep{pmlr-v119-tay20a}, Performer~\citep{choromanski2021rethinking}, Synthesizer~\citep{pmlr-v139-tay21a}, Nystr{\"{o}}mformer~\citep{xiong2021Nystromformer}, Luna~\citep{ma2021luna}, FNet~\citep{lee-thorp-etal-2022-fnet}, cosFormer~\citep{zhen2022cosformer}, and Paramixer~\citep{9878955}.

As shown in Table~\ref{tab:lra_res}, Converter consistently surpasses all baseline models in all five tasks with the best classification accuracy: ListOps (60.38\%), Text (86.44\%), Retrieval (83.41\%), Image (61.02\%), and Pathfinder (88.43\%). Converter attains an average accuracy of 75.94\%, substantially outperforming the second-best model Luna-256 (61.95\%) by a margin of 14 percentage points. Notably, on the challenging ListOps task, Converter (60.38\%) surpasses the second-best performer Paramixer (39.71\%) by more than 20.57\%, demonstrating its superior capability in handling structured sequential data. Furthermore, for the Image classification task, Converter (61.02\%) significantly outperforms the runner-up Luna-256 (47.86\%), showcasing its exceptional ability in visual feature extraction. These experimental results confirm the comprehensive advantages that Converter demonstrates in processing long-sequence tasks.

\subsection{Long Document Classification}
This task aims to evaluate the capability of Converter in modeling complex long-term dependencies for NLP tasks. We utilized a publicly available dataset collected from arXiv~\citep{a11080109}. Following \citet{9878955}, we selected four document categories: cs.AI, cs.NE, math.AC, and math.GR, yielding a dataset of 11956 documents. The documents were encoded at the character level, and all comparative models employed zero padding to achieve uniform length. The dataset was partitioned into 60\% training, 20\% validation, and 20\% test sets. Similar to the LRA benchmark, we created two standardized versions of the dataset through truncation: LongDoc16K with sequences of 16384 tokens and LongDoc32K with sequences of 32768 tokens. We then evaluated Converter and various Transformer-based architectures on these datasets.

\begin{table}[!h]
\centering
\caption{Accuracy results (\%) on long document classification. The best result is in bold and the second best is underlined.}
\label{tab:longdoc}
\begin{tabular}{l|cc}
\toprule
\textbf{Model} & \textbf{LongDoc16K}~$\uparrow$ & \textbf{LongDoc32K}~$\uparrow$ \\
\midrule
Vanilla Transformer & 68.39 & 70.84 \\
Linformer (Layerwise) & 49.03 & 45.00 \\
Performer & 62.26 & 65.65 \\
Synthesizer (Dense) & 76.05 & \underline{77.02} \\
Nystr{\"{o}}mformer & 67.26 & 69.27 \\
FNet & 44.92 & 46.94 \\
cosFormer & 64.44 & 67.26 \\
Paramixer (Chord) & \underline{79.60} & 74.76 \\
\midrule
Converter & \textbf{81.77} & \textbf{82.34} \\
\bottomrule
\end{tabular}
\end{table}

Table~\ref{tab:longdoc} illustrates that Converter surpasses all baseline models. Notably, synthetic attention-based Transformer variants (Synthesizer, Paramixer, and Converter) provide better results than self-attention approximation-based Transformers (Linformer, Performer, and Nystr{\"{o}}mformer) and the vanilla Transformer. Meanwhile, FNet, a parameter-free and attention-free attention-based Transformer variant, performs poorly in long document classification tasks. This demonstrates that high-rank or full-rank attention plays a crucial role in modeling long-term dependencies.

\subsection{DNA Sequence-based Taxonomy Classification}
We evaluated Converter against other models using biological data. We obtained cDNA sequences and their taxonomic labels from Ensembl~\footnote{\href{https://www.ensembl.org/index.html}{https://www.ensembl.org/index.html}} and designed two binary classification tasks. Similar to the long document classification task, each dataset was truncated to a fixed length of 16384 and partitioned into 60\% training, 20\% validation, and 20\% test sets. The first dataset, Ensembl (B/S), focuses on vertebrate organisms, comparing sequences from the genera Bos and Sus. While the classes are nearly balanced (51973 Bos and 50027 Sus sequences), this dataset is particularly challenging due to extreme variations in sequence length (ranging from 63 to 447010 bases). The second dataset, Ensembl (M/R), represents our most computationally intensive classification task at the genus level. This dataset compares genes from Mus and Rattus, featuring significant class imbalance with a nearly 2:1 ratio (275636 Mus and 133310 Rattus sequences). Sequence lengths vary substantially, spanning from 32 to 261093 bases.

\begin{table}[!h]
\centering
\caption{Accuracy results (\%) on DNA sequence-based taxonomy classification. The best result is in bold and the second best is underlined.}
\label{tab:dna}
\begin{tabular}{l|cc}
\toprule
\textbf{Model} & \textbf{Ensembl (B/S)}~$\uparrow$ & \textbf{Ensembl (M/R)}~$\uparrow$ \\
\midrule
Vanilla Transformer & 66.71 & 58.50 \\
Linformer (Layerwise) & 62.76 & 51.63 \\
Performer & 63.18 & 55.16 \\
Synthesizer (Dense) & 66.07 & 56.34 \\
Nystr{\"{o}}mformer & 66.15 & \underline{58.51} \\
FNet & 65.70 & 56.30 \\
cosFormer & 65.73 & 56.30 \\
Paramixer (Chord) & \underline{66.77}  & 56.37 \\
\midrule
Converter & \textbf{84.59} & \textbf{59.49} \\
\bottomrule
\end{tabular}
\end{table}

As shown in Table~\ref{tab:dna}, our model achieves strong performance. In Ensembl (B/S), Converter is 17.82\% more accurate than Paramixer. In Ensembl (M/R), Converter achieves an accuracy that is 0.98\% higher than Nystr{\"{o}}mformer. Moreover, our model consistently surpasses the vanilla Transformer in all two tasks.

\subsection{Ablation Studies}
We study the influence of different mechanisms used in Converter by ablating the corresponding components. Table~\ref{tab:lra_as} records the experimental results of Converter equipped with distinct components on the Long-Range Arena benchmark. NoPE means without position embedding, APE means learnable absolute position embedding~\citep{pmlr-v70-gehring17a}, and SPE means sinusoidal position embedding~\citep{NIPS2017_3f5ee243}. The ablation studies highlight the importance of RPE~\citep{neishi-yoshinaga-2019-relation}. Notably, Converter achieves the highest performance when using PRE, followed by APE and SPE in descending order, while the NoPE variant yields the lowest results. This finding contradicts recent papers regarding NoPE~\citep{haviv-etal-2022-transformer,chi-etal-2023-latent,NEURIPS2023_4e85362c}. Both versions of Converter with Kernolution (with and without the kernel polynomial loss) outperform Converter with Synvolution in all tasks.

\begin{table}[!h]
\centering
\caption{Ablation studies of Converter on the LRA benchmark.}
\label{tab:lra_as}
\begin{tabular}{l|ccccc}
\toprule
\textbf{Method} & \textbf{ListOps}~$\uparrow$ & \textbf{Text}~$\uparrow$ & \textbf{Retrieval}~$\uparrow$ & \textbf{Image}~$\uparrow$ & \textbf{Pathfinder}~$\uparrow$ \\
\midrule
Converter & 60.38 & 86.44 & 83.41 & 61.02 & 88.43 \\
\midrule
w/ NoPE & 36.44 & 62.31 & 66.85 & 41.01 & 77.48 \\
w/ SPE & 37.45 & 71.15 & 79.52 & 46.89 & 80.55 \\
w/ APE & 39.80 & 79.20 & 79.82 & 48.38 & 80.89 \\
w/ Synv. & 57.96 & 82.89 & 82.39 & 58.23 & 87.29 \\ 
w/o KPL & 59.32 & 83.60 & 83.11 & 59.75 & 88.18 \\
\bottomrule
\end{tabular}
\end{table}
\section{Conclusion and Future Work}
In this work, we introduce Converter, a Transformer variant that replaces self-attention with a synthetic unitary digraph convolution called Synvolution. By leveraging the inverse process of eigendecomposition and LQ factorization, we synthesize a unitary digraph shift operator with learnable eigenvalues and eigenvectors. Our fast $1$-DHHP implementation achieves linearithmic time complexity while preserving the full-rank property of Synvolution. We further enhance Synvolution by incorporating the kernel polynomial method, resulting in Kernelution. We propose a kernel polynomial loss to enable dynamic kernel adaptation during training.

We evaluate Converter on the Long-Range Arena benchmark, long document classification, and DNA sequence-based taxonomy classification tasks. The experimental results demonstrate the strong performance of Converter. This work takes a solid step forward in applying digraph convolution to large datasets under the architecture of Transformer. Future work will focus on developing the decoder component for cross-attention. We believe our synthetic unitary digraph convolution approach will inspire further research into the relationships among self-attention, digraph convolution, and convolution, opening new possibilities for designing more powerful and efficient neural network architectures that combine the strengths of these different approaches.

\newpage
\bibliographystyle{unsrtnat}
\bibliography{references}

\newpage
\appendix
\section{Pseudocode for Synvolution and Kernelution}
\begin{algorithm}[!h]
\caption{PyTorch-like pseudocode for Parallel Scan.}
\label{alg:pscan}
\definecolor{codeblue}{rgb}{0.25,0.5,0.5}
\lstset{
	backgroundcolor=\color{white},
	basicstyle=\fontsize{7.2pt}{7.2pt}\ttfamily\selectfont,
	columns=fullflexible,
	breaklines=true,
	captionpos=b,
	commentstyle=\fontsize{7.2pt}{7.2pt}\color{codeblue},
	keywordstyle=\fontsize{7.2pt}{7.2pt},
}
\begin{lstlisting}[language=python]
# b: batch size, n: length, d: feature dimension

class PScan(torch.autograd.Function):
    @staticmethod
    def expand_(A, X):
        if A.size(1) == 1:
            return
        T = 2 * (A.size(1) // 2)
        Aa = A[:, :T].view(A.size(0), T//2, 2, -1)
        Xa = X[:, :T].view(X.size(0), T//2, 2, -1)
        Xa[:, :, 1].add_(Aa[:, :, 1] * Xa[:, :, 0])
        Aa[:, :, 1].mul_(Aa[:, :, 0])
        PScan.expand_(Aa[:, :, 1], Xa[:, :, 1])
        Xa[:, 1:, 0].add_(Aa[:, 1:, 0] * Xa[:, :-1, 1])
        Aa[:, 1:, 0].mul_(Aa[:, :-1, 1])
        if T < A.size(1):
            X[:, -1].add_(A[:, -1] * X[:, -2])
            A[:, -1].mul_(A[:, -2])

    @staticmethod
    def acc_rev_(A, X):
        if X.size(1) == 1:
            return
        T = 2 * (X.size(1) // 2)
        Aa = A[:, -T:].view(A.size(0), T//2, 2, -1)
        Xa = X[:, -T:].view(X.size(0), T//2, 2, -1)
        Xa[:, :, 0].add_(Aa[:, :, 1].conj() * Xa[:, :, 1])
        B = Aa[:, :, 0].clone()
        B[:, 1:].mul_(Aa[:, :-1, 1].conj())
        PScan.acc_rev_(B, Xa[:, :, 0])
        Xa[:, :-1, 1].add_(Aa[:, 1:, 0].conj() * Xa[:, 1:, 0])
        if T < A.size(1):
            X[:, 0].add_(A[:, 1].conj() * X[:, 1])

    @staticmethod
    def forward(ctx, A, X, Y_init):
        ctx.A = A[:, :, None].clone()
        ctx.Y_init = Y_init[:, None, :].clone()
        ctx.A_star = ctx.A.clone()
        ctx.X_star = X.clone()
        PScan.expand_(ctx.A_star, ctx.X_star)
        return ctx.A_star * ctx.Y_init + ctx.X_star

    @staticmethod
    def backward(ctx, grad_output):
        U = grad_output * ctx.A_star.conj()
        A = ctx.A.clone()
        R = grad_output.clone()
        PScan.acc_rev_(A, R)
        Q = ctx.Y_init.expand_as(ctx.X_star).clone()
        Q[:, 1:].mul_(ctx.A_star[:, :-1].conj()).add_(ctx.X_star[:, :-1])
        grad_A = (Q.conj() * R).sum(-1)
        return grad_A, R, U.sum(dim=1)

pscan = PScan.apply
\end{lstlisting}
\end{algorithm}

Algorithm~\ref{alg:fast_order-1_dhhp_pytorch} presents the PyTorch-like pseudocode for $1$-DHHP as a discrete unitary transform and its inverse transform, while Algorithm~\ref{alg:kernelution} demonstrates the PyTorch-like pseudocode for Synvolution and Kernelution.

\begin{algorithm}[!h]
\caption{PyTorch-like pseudocode for $1$-DHHP as a discrete unitary transform and its inverse transform.}
\label{alg:fast_order-1_dhhp_pytorch}
\definecolor{codeblue}{rgb}{0.25,0.5,0.5}
\lstset{
	backgroundcolor=\color{white},
	basicstyle=\fontsize{7.2pt}{7.2pt}\ttfamily\selectfont,
	columns=fullflexible,
	breaklines=true,
	captionpos=b,
	commentstyle=\fontsize{7.2pt}{7.2pt}\color{codeblue},
	keywordstyle=\fontsize{7.2pt}{7.2pt},
}
\begin{lstlisting}[language=python]
# b: batch size, n: length, d: feature dimension
# m: permutation mapping dimension

def dhhp_trans(transform=True, x, g_l_ii, g_l_ij, g_l_ji, g_l_jj, g_u_ii, g_u_ij, g_u_ji, g_u_jj, diag):
    y = torch.zeros_like(x)
    z = torch.zeros_like(x)

    if transform is True:
        x = x.reshape(b, n // m, m, d).transpose(1, 2).reshape(b, n, d)
    else:
        x = torch.einsum('bn,bnd->bnd', diag, x)
    
    x_, y = torch.zeros_like(x), torch.zeros_like(x)
    x_[:, :-1, :] = g_u_ii.unsqueeze(-1) * x[:, :-1, :]
    x_ = x_.flip(1)
    g_u_ij = g_u_ij.flip(1)
    g_u_ij = torch.cat([g_u_ij, torch.zeros_like(g_u_ij[:, :1])], dim=1)
    p_u = torch.zeros_like(g_u_ij)
    p_u[:, 1:] = g_u_ij[:, :-1].clone()
    h_u_init = x_[:, 0, :].clone()
    h_u = pscan(p_u, x_, h_u_init)
    h_u = h_u.flip(1)
    y[:, 1:, :] = g_u_ji.unsqueeze(-1) * x[:, :-1, :] + g_u_jj.unsqueeze(-1) * h_u[:, 1:, :]
    y[:, 0, :] = h_u[:, 0, :]

    y_, z = torch.zeros_like(y), torch.zeros_like(y)
    y_[:, 1:, :] = g_l_jj.unsqueeze(-1) * y[:, 1:, :]
    g_l_ji = torch.cat([g_l_ji, torch.zeros_like(g_l_ji[:, :1])], dim=1)
    p_l = torch.zeros_like(g_l_ji)
    p_l[:, 1:] = g_l_ji[:, :-1].clone()
    h_l_init = y_[:, 0, :].clone()
    h_l = pscan(p_l, y_, h_l_init)
    z[:, :-1, :] = g_l_ii.unsqueeze(-1) * h_l[:, :-1, :] + g_l_ij.unsqueeze(-1) * y[:, 1:, :]
    z[:, n-1, :] = h_l[:, n-1, :]

    if transform is True:
        z = torch.einsum('bn,bnd->bnd', diag, z)
    else:
        z = z.reshape(b, m, n // m, d).transpose(1, 2).reshape(b, n, d)

    return z

def inverse_dhhp_trans(transform=False, x, g_l_ii_conj_trs, g_l_ij_conj_trs, g_l_ji_conj_trs, g_l_jj_conj_trs, g_u_ii_conj_trs, g_u_ij_conj_trs, g_u_ji_conj_trs, g_u_jj_conj_trs, diag_conj_trs):
    return dhhp_trans(transform, x, g_u_ii_conj_trs, g_u_ij_conj_trs, g_u_ji_conj_trs, g_u_jj_conj_trs, g_l_ii_conj_trs, g_l_ij_conj_trs, g_l_ji_conj_trs, g_l_jj_conj_trs, diag_conj_trs)
\end{lstlisting}
\end{algorithm}

\begin{algorithm}[!h]
\caption{PyTorch-like pseudocode for Synvolution and Kernelution.}
\label{alg:kernelution}
\definecolor{codeblue}{rgb}{0.25,0.5,0.5}
\lstset{
	backgroundcolor=\color{white},
	basicstyle=\fontsize{7.2pt}{7.2pt}\ttfamily\selectfont,
	columns=fullflexible,
	breaklines=true,
	captionpos=b,
	commentstyle=\fontsize{7.2pt}{7.2pt}\color{codeblue},
	keywordstyle=\fontsize{7.2pt}{7.2pt},
}
\begin{lstlisting}[language=python]
# b: batch size, n: length, d: feature dimension

def kernel_polynomial_method(seq, K, g, mu):
    Tx_0 = torch.ones_like(seq) # b, n
    cheb_gibbs = Tx_0 * mu[0]
    if K == 0:
        return cheb_gibbs
    
    Tx_1 = seq
    cheb_gibbs = cheb_gibbs + Tx_1 * g[1] * mu[1]
    if K == 1:
        return cheb_gibbs

    if K >= 2:
        for k in range(2, K+1):
        Tx_2 = 2 * seq * Tx_1 - Tx_0
        cheb_gibbs = cheb_gibbs + Tx_2 * g[k] * mu[k]
        Tx_0, Tx_1 = Tx_1, Tx_2

    return cheb_gibbs

def givens_rot_para(alpha, beta, gamma):
    g_ii = torch.exp(-1j * (alpha + beta) / 2) * torch.cos(gamma / 2)
    g_ij = -torch.exp(1j * (alpha - beta) / 2) * torch.sin(gamma / 2)
    g_ji = torch.exp(-1j * (alpha - beta) / 2) * torch.sin(gamma / 2)
    g_jj = torch.exp(1j * (alpha + beta) / 2) * torch.cos(gamma / 2)

    return g_ii, g_ij, g_ji, g_jj

def givens_rot_para_conj_trs(g_ii, g_ij, g_ji, g_jj):
    g_ii_conj_trs = g_jj
    g_ij_conj_trs = -g_ij
    g_ji_conj_trs = -g_ji
    g_jj_conj_trs = g_ii

    return g_ii_conj_trs, g_ij_conj_trs, g_ji_conj_trs, g_jj_conj_trs

def kernelution(eigenvalue, K, g, mu, x, alpha_l, beta_l, gamma_l, alpha_u, beta_u, gamma_u, theta):
    if enable_kernelution is True:
        eigenvalue = kernel_polynomial_method(eigenvalue, K, g, mu)

    g_l_ii, g_l_ij, g_l_ji, g_l_jj = givens_rot_para(alpha_l, beta_l, gamma_l)
    g_u_ii, g_u_ij, g_u_ji, g_u_jj = givens_rot_para(alpha_u, beta_u, gamma_u)
    diag = torch.exp(2j * math.pi * theta)
    
    g_l_ii_conj_trs, g_l_ij_conj_trs, g_l_ji_conj_trs, g_l_jj_conj_trs = givens_rot_para_conj_trs(g_l_ii, g_l_ij, g_l_ji, g_l_jj)
    g_u_ii_conj_trs, g_u_ij_conj_trs, g_u_ji_conj_trs, g_u_jj_conj_trs = givens_rot_para_conj_trs(g_u_ii, g_u_ij, g_u_ji, g_u_jj)
    diag_conj_trs = diag.conj()

    x = dhhp_trans(True, x, g_l_ii, g_l_ij, g_l_ji, g_l_jj, g_u_ii, g_u_ij, g_u_ji, g_u_jj, diag)
    x = torch.einsum('bn,bnd->bnd', eigenvalue, x)
    x = inverse_dhhp_trans(False, x, g_l_ii_conj_trs, g_l_ij_conj_trs, g_l_ji_conj_trs, g_l_jj_conj_trs, g_u_ii_conj_trs, g_u_ij_conj_trs, g_u_ji_conj_trs, g_u_jj_conj_trs, diag_conj_trs)
    
    return x
\end{lstlisting}
\end{algorithm}

\section{Synvolution and Other Attention Mechanisms}
We compare Synvolution with common attention and attention-like mechanisms in Table~\ref{tab:attention}. Notice, we do not assume the relation between the length size $N$ and embedded size $D$ for input signal $\mathbf{X} \in \mathbb{R}^{N \times D}$. Sparse attention, which reduces computation by only attending to a subset of tokens, includes \citep{child2019generating,NEURIPS2020_c8512d14}. Following the categorization in \citep{cao2021choose}, linear attention can be classified into Fourier-type and Galerkin-type. Fourier-type attention, which compute query-key pairs first, includes \citep{Shen_2021_WACV,cao2021choose,zhang-etal-2021-sparse,Koohpayegani_2024_WACV}. Galerkin-type attention, compute key-value pairs first, includes \citep{pmlr-v119-katharopoulos20a,cao2021choose,Koohpayegani_2024_WACV}. Kernel attention, which leverages kernel methods to approximate the attention operation, includes \citep{tsai-etal-2019-transformer,choromanski2021rethinking,NEURIPS2021_10a7cdd9,xiong2021Nystromformer}. Synthetic dense attention includes \citep{pmlr-v139-tay21a}. Chord attention includes \citep{9878955}. 

While sparse attention theoretically offers lower time complexity, its practical implementation remains challenging. When implemented directly in PyTorch without CUDA optimization~\citep{10.1145/1365490.1365500} or other GPU frameworks, matrix multiplication between sparse and dense tensors paradoxically consumes more GPU memory than multiplication between two dense tensors of equivalent size.

Linear attention mechanisms, whether Fourier-type or Galerkin-type, achieve lower time complexity compared to scaled dot-product attention. However, their attention matrices remain low-rank, similar to self-attention. For kernel attention mechanisms, their computational characteristics must be analyzed on a case-by-case basis due to implementation variations.

Synthesizer demonstrates potential through synthetic dense attention, though its time and space complexity remains equivalent to scaled dot-product attention. Chord attention, derived from the inverse process of Chord factorization~\citep{KHALITOV2022160}, lacks CUDA optimization in its implementation. This leads to higher GPU memory consumption than linear attention but lower than scaled dot-product attention, similar to the challenges faced by sparse attention implementations.

Our proposed method, Synvolution, is an attention-like mechanism that leverages a space-for-time approach. Its attention matrix possesses several advantageous properties: it is learnable, full-rank, dense, and unitary. These characteristics enable Synvolution to achieve effectiveness comparable to self-attention across diverse domains.

\begin{table*}[!h]
\centering
\caption{Overview of common attention and attention-like mechanisms.}\label{tab:attention}
\begin{tabular}{lccc}
\toprule
Method & Attention Matrix & Time Complexity & Space Complexity \\
\midrule
Scaled Dot-Product Attention & Learnable, Low-Rank, and Dense & $\mathcal{O}(N^{2}D + ND^{2})$ & $\mathcal{O}(N^{2} + ND)$ \\
Sparse Attention & Learnable, High/Full-Rank, and Sparse & $\mathcal{O}(|\mathcal{E}|D + ND^{2})$ & $\mathcal{O}(|\mathcal{E}| + ND)$ \\
Fourier-Type Attention & Learnable, Low-Rank, and Dense & $\mathcal{O}(N^{2}D + ND^{2})$ & $\mathcal{O}(N^{2} + ND)$ \\
Galerkin-Type Attention & Learnable, Rank-dependent, and Dense & $\mathcal{O}(ND^{2})$ & $\mathcal{O}(D^{2} + ND)$ \\
Kernelized Attention & Learnable, Low-Rank, and Dense & Depends on kernel & Depends on kernel \\
Synthetic Dense Attention & Learnable, Rank-Dependent, and Dense & $\mathcal{O}(N^{2}D + ND^{2})$ & $\mathcal{O}(N^{2} + ND)$ \\
Chord Attention & Learnable, Full-Rank, and Sparse & $\mathcal{O}(ND\log^{2}{N})$ & $\mathcal{O}(N\log^{2}{N} + ND)$ \\
\midrule
Synvolution & Learnable, Full-Rank, and Dense & $\mathcal{O}(ND\log{N}+ND^{2})$ & $\mathcal{O}(ND)$ \\
\bottomrule
\end{tabular}
\end{table*}

\section{Kernel Functions in Kernel Polynomial Method}
\begin{figure}[h]
\begin{center}
\includegraphics[scale=0.45]{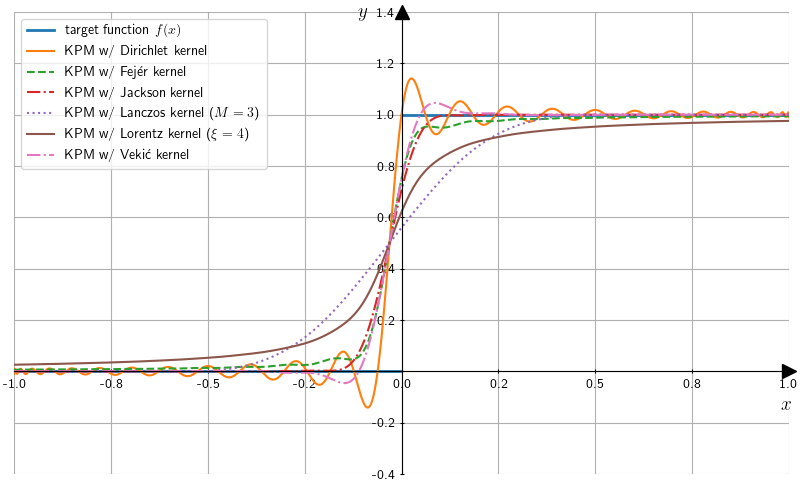}
\end{center}
\caption{An illustration of Gibbs phenomenon when using the kernel polynomial method with different kernels to approximate a step function.}
\label{fig:kpm}
\end{figure}

In Table~\ref{tab:kernel_func}, we summarize all currently known kernel functions used in the kernel polynomial method. To illustrate the approximation capabilities of different kernels when the Gibbs phenomenon occurs, we present a comparison in Figure~\ref{fig:kpm}, where the target function $f(x)$ is a sign step function. For more details about the kernel polynomial method, see \citet{RevModPhys.78.275} and \citet{Weiße2008}.

\begin{table}[!h]
\centering
\caption{Overview of kernel functions}\label{tab:kernel_func}
\resizebox{\textwidth}{!}{
\begin{tabular}{lccc}
\toprule
Kernel & Gibbs damping factor $g_k$ & Hyperparameters & Remarks \\
\midrule
Dirichlet & 1 & None & least favorable choice \\ 
Fej{\'e}r~\citep{Fejer1904} & $1-\frac{k}{K+1}$ & None & mainly of academic interest \\ 
Jackson~\citep{RevModPhys.78.275} & $\frac{(K+2-k)\cos(\frac{k\pi}{K+2})+\sin(\frac{k\pi}{K+2})\cot(\frac{\pi}{K+2})}{K+2}$ & None & optimal for most applications, but lacked rigorous proof \\ 
Lanczos~\citep{alma990004236840205776} & $\sinc^{M}(\frac{k\pi}{K+1})$ & $M \in \mathbb{N}$ & $M = 3$ closely matches the Jackson kernel \\ 
Lorentz~\citep{Vijay2004} & $\frac{\sinh\left[\xi(1-\frac{k}{K+1})\right]}{\sinh(\xi)}$ & $\xi \in \mathbb{R}$ & optimal for Green functions \\ 
Veki{\'c}~\citep{PhysRevLett.71.4283} & $\frac{1}{2} - \frac{1}{2}{\tanh}\left[\frac{\frac{k}{K+1}-\frac{1}{2}}{\frac{k}{K+1}(1-\frac{k}{K+1})}\right]$ & None & found empirically \\ 
Wang~\citep{PhysRevB.49.10154} & $\eu^{-\left(\frac{ak}{K+1}\right)^{b}}$ & $a, b \in \mathbb{R}$ & found empirically \\
\bottomrule
\end{tabular}}
\end{table}

\section{Kernel Polynomial Loss}
Here is the complete derivation process for Equation~\ref{eq:kpl}.
\begin{equation}
\centering
\begin{split}
\int_{-1}^{1}\left|\frac{\diff{f(x)}}{\diff{x}}\right|^{2}\diff{x}
&\approx \int_{-1}^{1}\left|\frac{\diff}{\diff{x}}\sum_{k=0}^{K}\mu_{k}T_k(x)\right|^{2}\diff{x} \\
&= \int_{-1}^{1}\left|\sum_{k=1}^{K}\mu_{k}T'_k(x)\right|^{2}\diff{x} \\
&= \int_{-1}^{1}\left(\sum_{k=1}^{K}\mu_{k}T'_k(x)\right)\left(\sum_{m=1}^{K}\overline{\mu_{m}}T'_m(x)\right)\diff{x} \\
&= \sum_{k=1}^{K}\sum_{m=1}^{K}\mu_{k}\overline{\mu_{m}}\int_{-1}^{1}T'_k(x)T'_m(x)\diff{x} \\
&= \sum_{k=1}^{K}|\mu_{k}|^{2}\int_{-1}^{1}|T'_k(x)|^{2}\diff{x} \\
&= \sum_{k=1}^{K}\pi k^{2}|\mu_{k}|^{2}
\end{split}
\end{equation}

\section{Proofs}
\noindent\textbf{Proof of Proposition~\ref{prop:dhhp_universal}}
\begin{proof}
First, we note that DFT, DWHT, DCT, and DST are all discrete unitary transforms, meaning their matrices are unitary. By Assumption~\ref{asmp:dhhp_givens_num_upper_bound}, any $N \times N$ unitary matrix can be exactly constructed using $L$-DHHP where $1 \leq L \leq \lceil\frac{N}{4}\rceil$. Therefore, as specific cases of unitary matrices, DFT, DWHT, DCT, and DST can all be exactly represented by $L$-DHHP. For their inverses, since the inverse of a unitary matrix is its conjugate transpose, they can also be exactly represented by $L$-DHHP.
\end{proof}

\noindent\textbf{Proof of Proposition~\ref{prop:dhhp_time_complexity}}
\begin{proof}
According to the definition of $L$-DHHP, the transform matrix $\bm{\Phi}$ can be decomposed as $\bm{\Phi} = \mathbf{D}\left(\prod_{l=1}^{L}\mathbf{H}^{(l)}_{\text{l}}\mathbf{H}^{(l)}_{\text{u}}\mathbf{P}^{(l)}\right)$, where $\mathbf{D}$ is a unitary diagonal matrix, $\mathbf{H}^{(l)}_{\text{l}}$ is a lower unitary Hessenberg matrix, $\mathbf{H}^{(l)}_{\text{u}}$ is a upper unitary Hessenberg matrix, and $\mathbf{P}^{(l)}$ is a permutation matrix. The computation of $\mathbf{y} = \bm{\Phi}\mathbf{x}$ can be broken down into sequential steps:
\begin{enumerate}
    \item For each order $l$ from $1$ to $L$:
        \item Multiplication with $\mathbf{P}^{(l)}$ requires $\mathcal{O}(N)$ operations
        \item Multiplication with $\mathbf{H}^{(l)}_{\text{u}}$ requires $\mathcal{O}(N\log{N})$ operations
        \item Multiplication with $\mathbf{H}^{(l)}_{\text{l}}$ requires $\mathcal{O}(N\log{N})$ operations
    \item The multiplication of diagonal matrix $\mathbf{D}$ with input signal $\mathbf{x}$ requires $\mathcal{O}(N)$ operations.
\end{enumerate}

As shown in Algorithm~\ref{alg:fast_order-1_dhhp_pytorch}, these multiplications can be implemented efficiently through Hadamard products, the total time complexity is $\mathcal{O}(LN\log{N})$.
\end{proof}

\noindent\textbf{Proof of Proposition~\ref{prop:dhhp_full_rank}}
\begin{proof}
($\Leftarrow$) First, we prove that if $\mathbf{D}$ is unitary, then $L$-DHHP is full-rank. By definition, every lower unitary Hessenberg matrix $\mathbf{H}^{(l)}_{\text{l}}$, upper unitary Hessenberg matrix $\mathbf{H}^{(l)}_{\text{u}}$, and permutation matrix $\mathbf{P}^{(l)}$ is unitary. Since the product of unitary matrices is unitary, and $\mathbf{D}$ is given to be unitary, the entire $L$-DHHP matrix is unitary. As every unitary matrix is full-rank, $L$-DHHP is full-rank.

($\Rightarrow$) Now we prove that if $L$-DHHP is full-rank, then $\mathbf{D}$ must be unitary. Let $\bm{\Phi} = \mathbf{D}\left(\prod_{l=1}^{L}\mathbf{H}^{(l)}_{\text{l}}\mathbf{H}^{(l)}_{\text{u}}\mathbf{P}^{(l)}\right)$ be $L$-DHHP. Since $\mathbf{H}^{(l)}_{\text{l}}$, $\mathbf{H}^{(l)}_{\text{u}}$, and $\mathbf{P}^{(l)}$ are all unitary, their product is unitary and thus full-rank. For $\bm{\Phi}$ to be full-rank, $\mathbf{D}$ must also be full-rank.
As $\mathbf{D}$ is diagonal, it is full-rank if and only if all its diagonal entries have unit magnitude, which is equivalent to $\mathbf{D}$ being unitary.
\end{proof}

\noindent\textbf{Proof of Theorem~\ref{theorem:cpi_diff}}
\begin{proof}
The complete proof can be found in \citet{10.1137/1.9781611975949}.
\end{proof}

\noindent\textbf{Proof of Theorem~\ref{theorem:cpi_ana}}
\begin{proof}
The complete proof can be found in \citet{10.1137/1.9781611975949}.
\end{proof}

\section{Additional Experimental Details}
\noindent\textbf{Baseline Implementations.} We use the PyTorch implementation released by the authors or the third part for all baseline models. We list all baseline model implementations as following.
\begin{itemize}
\item \textbf{Transformer}: \href{https://github.com/hyunwoongko/transformer}{https://github.com/hyunwoongko/transformer}
\item \textbf{Linformer}: \href{https://github.com/lucidrains/linformer}{https://github.com/lucidrains/linformer}
\item \textbf{Performer}: \href{https://github.com/lucidrains/performer-pytorch}{https://github.com/lucidrains/performer-pytorch}
\item \textbf{Synthesizer}: \href{https://github.com/10-zin/Synthesizer}{https://github.com/10-zin/Synthesizer}
\item \textbf{Nystr{\"{o}}mformer}: \href{https://github.com/mlpen/Nystromformer}{https://github.com/mlpen/Nystromformer}
\item \textbf{FNet}: \href{https://github.com/erksch/fnet-pytorch}{https://github.com/erksch/fnet-pytorch}
\item \textbf{cosFormer}: \href{https://github.com/OpenNLPLab/cosFormer}{https://github.com/OpenNLPLab/cosFormer}
\item \textbf{Paramixer}: \href{https://github.com/wiedersehne/Paramixer}{https://github.com/wiedersehne/Paramixer}
\end{itemize}

\subsection{Experimental Settings}
We provide the complete set of hyperparameters used to train all models and report their results. The optimal hyperparameters were determined using Bayesian optimization under a 16 GB GPU memory constraint. After selecting the best hyperparameters based on minimum validation loss, we evaluated the corresponding model on the test set and reported its performance. For all four datasets and seven baseline models, we employed the AdamW optimizer with cross-entropy loss. The training configurations and hyperparameters for each neural network and dataset are summarized in Tables~\ref{tab:baseline_configs} and Table~\ref{tab:converter_configs}.

\begin{table}[!h]
\centering
\caption{The final baseline model hyperparameters used in experiments. Abbreviations: PE for position embedding, ES for embedding size, HS for hidden size, NB for number of blocks, NH for number of heads, Pool for pooling strategy, BS for batch size, LR for learning rate, WR for weight decay, and N/A indicates that the corresponding parameter is not present in the architecture.}
\label{tab:baseline_configs}
\resizebox{\textwidth}{!}{
\begin{tabular}{llcccccccccccc}
\hline
Dataset  & Model & PE & ES & HS & NB & NH & Pool & BS & LR & WD & PE Dropout Rate & Value Dropout Rate & FFN Dropout Rate \\
\hline
LongDoc16K      & Transformer   & RPE & 64 & 256 & 2 & 2 & MEAN & 2 & 0.0005 & 0.0001 & 0.1 & 0.1 & 0.1 \\
                & Linformer     & RPE & 128 & 512 & 2 & 2 & MEAN & 4 & 0.0002 & 0.0001 & 0.1 & 0.1 & 0.1 \\
                & Performer     & RPE & 128 & 512 & 2 & 2 & MEAN & 4 & 0.0002 & 0.0001 & 0.1 & 0.1 & 0.1 \\
                & Synthesizer   & RPE & 64 & 256 & 2 & 2 & MEAN & 2 & 0.0005 & 0.0001 & 0.1 & 0.1 & 0.1 \\
                & FNet          & RPE & 128 & 512 & 2 & N/A & MEAN & 4 & 0.0002 & 0.0001 & 0.1 & 0.1 & 0.1 \\
                & cosFormer     & RPE & 128 & 512 & 2 & 2 & MEAN & 4 & 0.0002 & 0.0001 & 0.1 & 0.1 & 0.1 \\
                & Paramixer     & RPE & 128 & 512 & 2 & N/A & MEAN & 4 & 0.0002 & 0.0001 & 0.1 & 0.1 & 0.1 \\
\hline
LongDoc16K      & Transformer   & RPE & 64 & 256 & 2 & 1 & MEAN & 2 & 0.0005 & 0.0001 & 0.1 & 0.1 & 0.1 \\
                & Linformer     & RPE & 128 & 512 & 2 & 2 & MEAN & 4 & 0.0002 & 0.0001 & 0.1 & 0.1 & 0.1 \\
                & Performer     & RPE & 128 & 512 & 2 & 2 & MEAN & 4 & 0.0002 & 0.0001 & 0.1 & 0.1 & 0.1 \\
                & Synthesizer   & RPE & 64 & 256 & 2 & 1 & MEAN & 2 & 0.0005 & 0.0001 & 0.1 & 0.1 & 0.1 \\
                & FNet          & RPE & 128 & 512 & 2 & N/A & MEAN & 4 & 0.0002 & 0.0001 & 0.1 & 0.1 & 0.1 \\
                & cosFormer     & RPE & 128 & 512 & 2 & 2 & MEAN & 4 & 0.0002 & 0.0001 & 0.1 & 0.1 & 0.1 \\
                & Paramixer     & RPE & 128 & 512 & 2 & N/A & MEAN & 4 & 0.0002 & 0.0001 & 0.1 & 0.1 & 0.1 \\
\hline
Ensembl (B/S)   & Transformer   & RPE & 64 & 256 & 2 & 1 & MEAN & 2 & 0.0002 & 0.0001 & 0.1 & 0.1 & 0.1 \\
                & Linformer     & RPE & 64 & 256 & 2 & 2 & MEAN & 2 & 0.0001 & 0.0001 & 0.1 & 0.1 & 0.1 \\
                & Performer     & RPE & 64 & 256 & 2 & 2 & MEAN & 2 & 0.0001 & 0.0001 & 0.1 & 0.1 & 0.1 \\
                & Synthesizer   & RPE & 64 & 256 & 2 & 1 & MEAN & 2 & 0.0002 & 0.0001 & 0.1 & 0.1 & 0.1 \\
                & FNet          & RPE & 64 & 256 & 2 & N/A & MEAN & 2 & 0.0001 & 0.0001 & 0.1 & 0.1 & 0.1 \\
                & cosFormer     & RPE & 64 & 256 & 2 & 2 & MEAN & 2 & 0.0001 & 0.0001 & 0.1 & 0.1 & 0.1 \\
                & Paramixer     & RPE & 64 & 256 & 2 & N/A & MEAN & 2 & 0.0001 & 0.0001 & 0.1 & 0.1 & 0.1 \\
\hline
Ensembl (M/R)   & Transformer   & RPE & 64 & 256 & 2 & 1 & MEAN & 2 & 0.0002 & 0.0001 & 0.1 & 0.1 & 0.1 \\
                & Linformer     & RPE & 64 & 256 & 2 & 2 & MEAN & 2 & 0.0001 & 0.0001 & 0.1 & 0.1 & 0.1 \\
                & Performer     & RPE & 64 & 256 & 2 & 2 & MEAN & 2 & 0.0001 & 0.0001 & 0.1 & 0.1 & 0.1 \\
                & Synthesizer   & RPE & 64 & 256 & 2 & 1 & MEAN & 2 & 0.0002 & 0.0001 & 0.1 & 0.1 & 0.1 \\
                & FNet          & RPE & 64 & 256 & 2 & N/A & MEAN & 2 & 0.0001 & 0.0001 & 0.1 & 0.1 & 0.1 \\
                & cosFormer     & RPE & 64 & 256 & 2 & 2 & MEAN & 2 & 0.0001 & 0.0001 & 0.1 & 0.1 & 0.1 \\
                & Paramixer     & RPE & 64 & 256 & 2 & N/A & MEAN & 2 & 0.0001 & 0.0001 & 0.1 & 0.1 & 0.1 \\
\hline
\end{tabular}}
\end{table}

\begin{table}[!h]
    \centering
    \caption{The hyperparameters of Converter for all experiments. Abbreviations: PE for position embedding, ES for embedding size, HS for hidden size, K for the maximum order of KPM, Pool for pooling strategy, BS for batch size, LR for learning rate, and WR for weight decay.}
    \resizebox{\textwidth}{!}{
    \begin{tabular}{l c c c c c c c c c c c c c c}
    \toprule
    Dataset  & PE & ES & HS & K & $\eta$ & Pool & BS & LR & WD & PE Dropout Rate & Value Dropout Rate & GFFN Dropout Rate & Eigenvalue Dropout Rate & Eigenvector Dropout Rate \\ 
    \midrule
    ListOps & RPE & 32 & 128 & 2 & 0.001 & MEAN & 128 & 0.001 & 0.001 & 0.1 & 0.1 & 0.1 & 0.1 & 0.1 \\
    Text & RPE & 64 & 256 & 2 & 0.001 & MEAN & 128 & 0.001 & 0.001 & 0.1 & 0.1 & 0.1 & 0.1 & 0.1 \\
    Retrieval & RPE & 64 & 256 & 2 & 0.001 & MEAN & 256 & 0.001 & 0.001 & 0.1 & 0.1 & 0.1 & 0.1 & 0.1 \\
    Image & RPE & 64 & 256 & 2 & 0.01 & MEAN & 128 & 0.001 & 0.001 & 0.1 & 0.1 & 0.1 & 0.1 & 0.1 \\
    Pathfinder & RPE & 64 & 256 & 2 & 0.001 & MEAN & 256 & 0.0002 & 0.0002 & 0.1 & 0.1 & 0.1 & 0.1 & 0.1 \\
    \midrule
    LongDoc16K & RPE & 128 & 512 & 2 & 0.1 & MEAN & 4 & 0.001 & 0.001 & 0.1 & 0.1 & 0.1 & 0.1 & 0.1 \\
    LongDoc32K & RPE & 128 & 512 & 2 & 0.1 & MEAN & 4 & 0.001 & 0.001 & 0.1 & 0.1 & 0.1 & 0.1 & 0.1 \\
    \midrule
    Ensembl (B/S) & RPE & 128 & 512 & 2 & 0.1 & MEAN & 32 & 0.001 & 0.001 & 0.1 & 0.1 & 0.1 & 0.1 & 0.1 \\
    Ensembl (M/R) & RPE & 128 & 512 & 2 & 0.1 & MEAN & 32 & 0.001 & 0.001 & 0.1 & 0.1 & 0.1 & 0.1 & 0.1 \\
    \bottomrule
    \end{tabular}}
    \label{tab:converter_configs}
\end{table}

\end{document}